\documentclass[submission,copyright]{eptcs}
\providecommand{\event}{ECAI AREA Workshop 2020} % Name of the event you are submitting to
\usepackage{breakurl}             % Not needed if you use pdflatex only.

\let\proof\relax
\let\endproof\relax

\usepackage[ruled,vlined,linesnumbered,noend]{algorithm2e}
\usepackage{amsbsy}
\usepackage{amsfonts}
\usepackage{amsmath}
\usepackage{amssymb}
\usepackage{amsthm}
\usepackage[small]{caption}
\usepackage{color}
\usepackage{epsfig}
\usepackage{flushend}
\usepackage{graphicx}
\usepackage{soul}
\usepackage{subcaption}

\graphicspath{figures/}

\newtheorem{proposition}{Proposition}
\newtheorem{definition}{Definition}

\newcommand{\R}{\mathbb{R}}

\newcommand{\mc}{\mathcal}
\newcommand{\ol}{\overline}
\newcommand{\suchthat}{\hspace{2mm} | \hspace{2mm}}

\DeclareMathOperator*{\argmax}{argmax}
\DeclareMathOperator*{\argmin}{argmin}

\newenvironment{proofsketch}{%
    \proof}{\endproof}

\title{\LARGE \bf Improving Competence for Reliable Autonomy}

\author{
	Connor Basich
	\institute{University of Massachusetts \\ Amherst, Massachusetts, USA}
	\email{cbasich@cs.umass.edu}
    \and
	Justin Svegliato
	\institute{University of Massachusetts \\ Amherst, Massachusetts, USA}
	\email{jsvegliato@cs.umass.edu}
    \and
    	Shlomo Zilberstein
	\institute{University of Massachusetts \\ Amherst, Massachusetts, USA}
	\email{shlomo@cs.umass.edu}
    \and
	Kyle Hollins Wray
	\institute{Alliance Innovation Lab Silicon Valley \\ Santa Clara, California, USA}
        \email{kyle.wray@nissan-usa.com}
    \and
	Stefan J. Witwicki
    	\institute{Alliance Innovation Lab Silicon Valley \\ Santa Clara, California, USA}
        \email{stefan.witwicki@nissan-usa.com}
}

\def\titlerunning{Improving Competence for Reliable Autonomy}
\def\authorrunning{C. Basich, J. Svegliato, K. Wray, S. Witwicki, and S. Zilberstein}

\begin{document}

\maketitle
\thispagestyle{empty}
\pagestyle{empty}

\begin{abstract}
    Given the complexity of real-world, unstructured domains, it is often impossible or impractical to design models that include every feature needed to handle all possible scenarios that an autonomous system may encounter. For an autonomous system to be reliable in such domains, it should have the ability to improve its competence online. In this paper, we propose a method for improving the competence of a system over the course of its deployment. We specifically focus on a class of semi-autonomous systems known as \emph{competence-aware systems} that model their own competence---the optimal extent of autonomy to use in any given situation---and learn this competence over time from feedback received through interactions with a human authority. Our method exploits such feedback to identify important state features missing from the system's initial model, and incorporates them into its state representation. The result is an agent that better predicts human involvement, leading to improvements in its competence and reliability, and as a result, its overall performance.
\end{abstract}

\section{Introduction}
\label{sec:intro}
Recent advances in artificial intelligence and robotics have enabled the deployment of autonomous systems in domains of increasing complexity and over long durations. Examples include autonomous navigation~\cite{broggi1999argo,dickmanns2007dynamic,broggi2012vislab}, extraterrestrial exploration~\cite{mustard2013mars,gao2017review}, and personal assistance~\cite{meeussen2011long,hawes2017strands}. However, despite substantial progress in these areas, it is still often either impossible or impractical to design models with every feature needed to handle all possible scenarios that may be encountered by an autonomous system~\cite{svegliato2019belief}. This can be due to the complexity of the state space, a lack of information, or unknown personal needs and preferences of stakeholders. As a result, many autonomous systems still rely on human assistance in various capacities to successfully accomplish their tasks. Ideally, to diminish their reliance on a human operator in unanticipated situations, an autonomous system should have the ability to identify and introduce missing features to their models during deployment, increasing the extent of their autonomous operation.

Competence-aware systems (CAS) have been recently proposed as a planning framework to reduce unnecessary reliance on human assistance~\cite{basich2020learning}. However, these systems operate on a fixed model, making it hard to successfully operate in situations where the system lacks key features in its state representation. In general, a CAS is a semi-autonomous system~\cite{wray2016hierarchical} that operates in, and plans for, different levels of autonomy, each of which corresponds to unique constraints on its capabilities and unique interactions with a human authority~\cite{basich2020learning}. For example, a service robot may request that the human operator supervise or even take control when an unrecognized obstacle blocks its way. Reliance on a human operator in these situation stems from limited competence on the part of the autonomous system. Competence-aware systems (CAS) were developed to provide a formal model for enabling the system to learn its competence, optimize its autonomy given its learned competence, and use this knowledge to improve the efficiency and reliability of its decision making while reducing unnecessary reliance on humans.

While the CAS model enables a semi-autonomous system to optimize its autonomy over time, it is still limited by the features in its fixed model. For example, consider a robot that is deployed on a campus with the task of delivering packages to various offices in different buildings. As a baseline, the robot can detect crosswalks and knows that it must ask for approval or supervision before crossing them. However, the robot only uses the number of vehicles in close proximity to reason about any crosswalk, while the human authority who approves or supervises the robot may use more detailed features, such as the visibility at the crosswalk, the time of day, or the weather conditions. Since the CAS model does not have the ability to add features to its state representation, the human feedback may appear inconsistent to the robot and lead to low competence, poor performance, and extra burden on the human.

In this paper, we propose a method for providing competence-aware systems the ability to improve their competence over time by increasing the granularity of the state representation through online model updates, leading to a more nuanced drawing of the boundaries between regions with different levels of competence. To do this, we leverage existing human feedback acquired through the agent's interactions with the human authority and identify instances where the feedback is inconsistent and appears random. As the system expects human feedback to be consistent up to small noise, instances where this expectation is not met are situations in which the agent's model is likely lacking some feature(s) that the human's feedback depends on. By adding these features to the model, the system can improve its predictive capabilities of the human's involvement, better modeling the value of operating in each level of autonomy. This leads to an increase in the system's competence and, consequently, the system's overall performance.

To validate our approach, we test both a standard CAS without the ability to modify its feature space and a modified CAS with the ability to modify its feature space, on a simulated domain where a robotic agent is tasked with delivering packages to various locations throughout the map. 
To complete its task, the agent must navigate multiple obstacles that it initially represents with only a few of the features used by the human authority. We show that the modified CAS correctly identifies missing causal features used by the human authority and adds them to its feature space, leading to a higher competence, a more accurate model, and a lower cost of operation compared to the standard CAS.

\section{Background on Competence Awareness}
\label{sec:background}
We begin by reviewing the primary model used in this approach. A competence-aware system (CAS) is a semi-autonomous system~\cite{wray2016hierarchical} that operates in and plans for multiple levels of autonomy, each of which is associated with different restrictions on autonomous operation and distinct forms of human involvement~\cite{basich2020learning}. A CAS combines three different models---a \emph{domain model}, an \emph{autonomy model}, and a \emph{human feedback model}---into a single decision making framework.

\subsection{Domain Model}

The \emph{domain model} (DM) models the environment in which the agent operates as a stochastic shortest path (SSP) problem. An SSP is a general-purpose model for sequential decision making in stochastic domains with an objective of finding the least-cost path from a start state to a goal state. This model has been used in a wide range of applications, including exception recovery~\cite{svegliato2019belief}, electric vehicle charging~\cite{saisubramanian2017optimizing}, search and rescue~\cite{pineda2015continual}, and autonomous navigation~\cite{wray2016hierarchical}.

Formally, an SSP is represented by the tuple $\langle S, A, T, C, s_0, s_g \rangle$, where $S$ is a finite set of states, $A$ is a finite set of actions, $T : S \times A \times S \rightarrow [0,1]$ represents the probability of reaching state $s' \in S$ after performing action $a \in A$ in state $s \in S$, $C : S \times A \rightarrow \mathbb{R}^+$ represents the expected immediate cost of performing action $a \in A$ in state $s \in S$, $s_0$ is a start state, and $s_g$ is a goal state such that $\forall a \in A, T(s_g, a, s_g) = 1 \land C(s_g, a) = 0$.

A solution to an SSP is a policy $\pi : S \rightarrow A$ that indicates that action $\pi(s) \in A$ should be taken in state $s \in S$. A policy $\pi$ induces the value function $V^{\pi} : S \rightarrow \mathbb{R}$ that represents the expected cumulative cost $V^{\pi}(s)$ of reaching the goal state $s_g$ from state $s$ following policy $\pi$. An optimal policy $\pi^*$ minimizes the expected cost of reaching the goal from the start state, $V^*(s_0)$.

\subsection{Autonomy Model}

The \emph{autonomy model} (AM) models the extent of autonomous operation the system can perform, i.e. the forms of operation and the external constraints imposed on when each form is allowed. $AM$ is represented by the tuple $\langle \mc{L}, \kappa, \mu \rangle$. $\mc{L} = \{ l_0, ..., l_n \}$ is the set of \emph{levels of autonomy}, where each $l_i$ corresponds to a set of constraints on the system's autonomous operation. Although not required, each level of autonomy generally involves a form of human involvement that reflects, or compensates for, the constraints imposed on the autonomy. For instance, it is common in autonomous service robots to have some level of \emph{supervised autonomy} which allows for autonomous operation conditioned on the requirement that a human is ready and available to monitor the system and override it if they deem an action to be unsafe or undesirable.
$\kappa: S \times A \rightarrow \mc{P}(\mc{L})$ is the \emph{autonomy profile} that indicates the allowed levels of autonomy when performing action $a \in A$ in state $s \in S$. $\kappa$ may model external constrains that could represent legal or ethical considerations.
%% , for example prohibiting an autonomous vehicle from operating autonomously in certain situations, such as when driving by a school. 
$\kappa$ constrains the full policy space so that the system can never follow a policy that violates $\kappa$.
$\mu: S \times \mc{L} \times A \times \mc{L} \rightarrow \R$ is the \emph{cost of autonomy} that represents the cost of performing action $a \in A$ at level $l' \in \mc{L}$ in state $s \in S$ having just operated in level $l \in \mc{L}$.

\subsection{Human Feedback Model}

The \emph{human feedback model} (HM) models the autonomous agent's current knowledge and belief about its interactions with the human agent. $HM$ is represented by the tuple $\langle \Sigma, \lambda, \rho, \tau \rangle$, where $\Sigma = \{ \sigma_0, ..., \sigma_n \}$ is the set of possible feedback signals the agent can receive from the human, $\lambda : S \times \mc{L} \times A \times \mc{L} \rightarrow \Delta^{|\Sigma|}$ is the \textit{feedback profile} that represents the probability of receiving signal $\sigma$ when performing action $a \in A$ at level $l' \in \mc{L}$ given that the agent is in state $s \in S$ and just operated in level $l \in \mc{L}$, $\rho : S \times \mc{L} \times A \times \mc{L} \rightarrow \R^+$ is the \textit{human cost function} that represents the cost to the human of performing action $a \in A$ at level $l' \in \mc{L}$ given that the agent is in state $s \in S$ and just operated in level $l \in \mc{L}$, and $\tau: S \times A \rightarrow \Delta^{|S|}$ is the \textit{human state transition function} that represents the probability of the human taking the agent to state $s' \in S$ when the agent attempted to perform action $a \in A$ in state $s \in S$ but the human took over control. In general, the human's true feedback profile, $\lambda^\mc{H}$, and transition function, $\tau^\mc{H}$, are likely unknown to the agent, which instead operates on some, possibly data-driven, estimate of each.

\subsection{Competence-Aware Systems}
\subsubsection*{Definition of a CAS}
A competence-aware system combines the three models defined above into one planning model by augmenting the base domain model with information from the autonomy model and the human feedback model. Solving a CAS then produces a policy that exploits its information about the human feedback to best utilize the levels of autonomy that are allowed by its autonomy profile. Formally, we define a CAS as the following augmented SSP:
\vspace{10pt}
\begin{definition}
    A \textbf{competence-aware system} $\mc{S}$ is represented by the tuple
    $\langle \ol{S}, \ol{A}, \ol{T}, \ol{C}, \ol{s}_0, \ol{s}_g \rangle$, where:
    \begin{itemize}
        \item $\ol{S} = S \times \mc{L}$ is a set of factored states, with each defined by a domain state and a level of autonomy.
        \item $\ol{A} = A \times \mc{L}$ is a set of factored actions, with each defined by a domain action and a level of autonomy.
        \item $\ol{T} : \ol{S} \times \ol{A} \rightarrow \Delta^{|\ol{S}|}$ is a transition function comprised of $T : S \times A \rightarrow \Delta^{|S|}$, $\lambda : \ol{S} \times \ol{A} \rightarrow \Delta^{|\Sigma|}$, and $\tau: S \times A \rightarrow \Delta^{|S|}$.
        \item $\ol{C} : \ol{S} \times \ol{A} \rightarrow \R^+$ is a cost function comprised of $C: S \times A \rightarrow \R^+$, $\mu : \ol{S} \times \ol{A} \rightarrow \R$, and $\rho : \ol{S} \times \ol{A} \rightarrow \R^+$.
        \item $\ol{s}_0 \in \ol{S}$ is the initial state $\ol{s}_0 = \langle s_0, l \rangle$ for some $l \in \mc{L}$.
        \item $\ol{s}_g \in \ol{S}$ is the goal state $\ol{s}_g = \langle s_g, l \rangle$ for some $l \in \mc{L}$.
    \end{itemize}
\end{definition}

\noindent The objective of a CAS is to find an optimal policy $\pi^*_\kappa \in \Pi$ that minimizes the value function $V^\pi$ subject to the condition that, for every state $\ol{s} = (s, l') \in \ol{S}$, the policy $\pi(\ol{s})$ \emph{never} indicates an action $\ol{a} = (a, l) \in A$ for which the level of autonomy $l$ is \emph{not} allowed for state $s$ and action $a$ by $\kappa(s, a)$.

\subsubsection*{Example of a CAS}
\label{sec:example}
To illustrate the role of the various functions, we describe the CAS model used in our experiments in Section~\ref{sec:experiments}, which is the same CAS defined in Section 4.2 of~\cite{basich2020learning}.

The CAS can operate in one of the four levels, $\mc{L} = \{l_0, l_1, l_2, l_3\}$, corresponding to (1) no autonomy, (2) verified autonomy, (3) supervised autonomy, and (4) unsupervised autonomy. Level 1 requires a human to complete the action for the agent, either through direct control, tele-operation, or simply manually performing the action themselves. Level 2 requires the agent to query for explicit approval from the human prior to executing the action. Level 3 requires a human to be available in a supervisory capacity while the agent executes the action, with the ability to override and take control of the system if deemed necessary. Level 4 requires no human involvement and is performed fully autonomously.

The system can receive the the following four feedback signals: approval ($\oplus$), disapproval ($\ominus$), override ($\oslash$), and no signal ($\emptyset$). We furthermore assume that approval and disapproval can only be received in $l_1$, and the system always receives a feedback signal in $l_1$, and can only receive $\oslash$ in $l_2$. 

We can now specify the \textbf{state transition function} of this CAS. Given 
$\ol{s}, \ol{a},$ and $\ol{s}'$, we define $\ol{T}$ as follows:
\begin{equation}
        \ol{T}(\ol{s}, \ol{a}, \ol{s}') = 
        \begin{cases}
            \tau(s, a, s'), & \text{if $l = l_0$},\\
            \lambda( \oplus ) T(s, a, s')+ \lambda( \ominus ) [s = s'], & \text{if $l = l_1$},\\
            \lambda( \emptyset ) T(s, a, s')+ \lambda( \oslash ) \tau(s, a, s'), & \text{if $l = l_2$},\\
            T(s, a, s'), & \text{if $l = l_3$},
        \end{cases}
        \label{eq: T-bar}
\end{equation}
where $\lambda(\cdot) = \lambda(\cdot | \ol{s}, \ol{a})$ and $[\cdot]$ denotes Iverson brackets.

Intuitively, in $l_0$, the system acts according to its estimate of the human's transition function. In $l_1$, the system operates according to its own transition when it receives approval, and stays in the same state when it receives disapproval. In $l_2$, the system acts according to its estimate of the human's transition function when it is overridden, and otherwise acts according to its own transition function. In $l_3$, it simply acts according to its own transition function. 

We define $\ol{C}$ as follows:
\begin{equation}
    \ol{C}(\ol{s}, \ol{a}) = g\big(C(s, a), \mu(\ol{s}, \ol{a}), \rho(\ol{s}, \ol{a})\big),
\end{equation}
where $g$ is any cost aggregation function on $C, \mu$, and $\rho$. The simplest case of which is a a linear combination of the three factors. 

\subsubsection*{Properties of a CAS}
We now review two key notions of a CAS: \emph{competence} and \emph{level-optimality}. First, the \emph{competence} of a CAS for executing action $a$ in state $\ol{s}$ is the most cost-effective level of autonomy given perfect knowledge of the human's feedback model. If the human is likely to deny or override the system autonomously carrying out action $a$, the competence will likely be low. Similarly, if the human is likely to allow the action to be carried out autonomously, we expect the competence to be high, although this is not formally required. It is worth emphasizing that this is a definition on the human-agent system as a whole, and not simply the underlying autonomous agent, as the competence is directly affected not just by the technical capabilities of the agent, but also the human's perception of the agent's capabilities.
\begin{definition}
    \label{def:lambda_h}
    Let $\lambda^{\mc{H}}$ be the stationary distribution of feedback signals that the human authority follows.  The \textbf{competence} of CAS $\mc{S}$, denoted $\chi_\mc{S}$, is a mapping from $\ol{S} \times A$ to the optimal (least-cost) level of autonomy given perfect knowledge of $\lambda^{\mc{H}}$. Formally:
    \begin{equation*}
        \chi_\mc{S}(\ol{s}, a) = \argmin_{l \in L} Q(\ol{s}, (a, l) ; \lambda^{\mc{H}})
    \end{equation*}
    where $Q(\ol{s}, (a, l) ; \lambda^{\mc{H}})$ is the expected cumulative reward when taking action $\ol{a} = (a, l)$ in state $\ol{s}$ conditioned on knowing the human's true feedback distribution, $\lambda^{\mc{H}}$.
\end{definition}

Second, we say that a CAS is \emph{level-optimal} in state $\ol{s}$ if the system operates at its competence in that state. In other words, it is a measure of how often the system operates at its competence. The higher the system's level-optimality is, the better it has learned to interact with and exploit the capabilities of the human authority as well as its own capabilities.
\begin{definition}
    A CAS $\mc{S}$ is \textbf{level-optimal} if
    \[
    \pi^*(\ol{s}) = (a, \chi_\mc{S}(\ol{s}, a)) \hspace{5mm} \forall \ol{s} \in \ol{S}
    \]
    Similarly, $\mc{S}$ is $\gamma$-\textbf{level-optimal} if this holds for an $\gamma$ portion of states. 
\end{definition}

\section{Improving Competence}
\label{sec:competence}
While our recent work on competence awareness proves that under a set of assumptions a CAS will converge to be level-optimal in the limit, this result says nothing about the quality of the CAS's competence~\cite{basich2020learning}. In particular, if a CAS is missing the features necessary to correctly represent its domain in a way that aligns with the human, even it converges to be level-optimal, its competence may be quite low.

Hence, the objective of this work is to \textbf{provide a CAS with the ability to improve its competence over time} by leveraging the existing human feedback available to the agent. Formally, this means that the CAS will increase both its competence in various situations and the total level-optimality of the system. Our approach relies on the assumption that humans are $\epsilon$-\emph{consistent} in their feedback; that is, given the same $(\ol{s}, \ol{a})$, the feedback signal returned will be the same up to some small noise $\epsilon$. We address practical concerns regarding this assumption in Section~\ref{sec:assumptions}. Under this assumption, the system can identify cases where feedback appears inconsistent or random, indicating a potential \emph{missing} feature---a feature used by the human authority to make their decisions, but currently not used by the system. By identifying the most likely feature or combination of features missing from the system's domain model, the agent can update its model to better align with the internal model of the human, enabling it to improve its overall competence by discriminating between situations where it can and cannot act in any given level of autonomy. 

\subsection{Definitions}

Before presenting the general algorithm of our approach, we begin with several important definitions. Let $\mc{S} = \langle AM, HM, DM \rangle$ be a competence-aware system. 

The \emph{complete feature space} available to $\mc{S}$, e.g. from its sensors or other external sources, can be partitioned into an \emph{active feature space} that is used by $\mc{S}$ and an \emph{inactive feature space} that is not yet used by $\mc{S}$. As $\mc{S}$ receives additional feedback over time, $\mc{S}$ will learn to exploit inactive features in order to more effectively align with the features used by the human authority.

\begin{definition}
    Given the \textbf{complete feature space} $F = F_1 \times F_2 \times \cdots \times F_n$ produced by the set of sensors available to $\mc{S}$, the \textbf{active feature space} used by $\mc{S}$ is $\hat{F} = \hat{F}_1 \times \cdots \times \hat{F}_m \subset F$ while the \textbf{inactive feature space} not used by $\mc{S}$ is $\breve{F} = F \setminus \hat{F}$.
\end{definition}

In order to ensure that level-optimality can be improved, we assume that the human authority produces feedback that remains \emph{consistent} during the operation of $\mc{S}$. This means that when the agent performs the same action in the same state at the same level of autonomy, it expects to receive, with high probability, the same feedback each time; observing consistent violation of this assumption is central to our approach.

\begin{definition}
    A human's feedback is \textbf{$\epsilon$-\emph{consistent}} if with probability at least $1 - \epsilon$, for some small noise $\epsilon$, the feedback provided for some $(\ol{s}, \ol{a})$ is the same each time $(\ol{s},\ol{a})$ is encountered. If this fails to hold for any sufficiently small $\epsilon$, the feedback is said to be \textbf{inconsistent}.
\end{definition}

We now present the two central concepts of our paper. (1) A state $\ol{s}$ is \emph{indiscriminate} if it is missing information, leading to the feedback profile having a low predictive confidence. Intuitively, the condition states that for at least one action, there is \emph{no} feedback signal that the system predicts with sufficiently high confidence. (2) A \emph{discriminator} is a feature (or set of features) in $\breve{F}$ that enables the feedback profile to better discriminate the feedback of an indiscriminate state. For example, consider a state that represents a closed door. With no additional features, the agent may have received, say, 50\% approvals and 50\% disapprovals. After adding the feature \emph{door size}, the agent may observe that it receives 100\% approval for light doors, and 100\% disapproval for heavy doors.

\begin{definition}
    We say that a state $\ol{s}$ is \textbf{indiscriminate} if it satisfies the following condition:
    \[
        \exists \hspace{1mm} \ol{a} \in \ol{A} \suchthat \forall \sigma \in \Sigma: \hat{\lambda}(\sigma | \ol{s}, \ol{a}) \leq 1 - \delta, \hspace{2mm} \delta \in \Big(\epsilon, 1 - \frac{1}{|\Sigma|}\Big)
    \]
    where the human is $\epsilon$-consistent.
    \label{def:indiscriminate_state}
\end{definition}
Note that $\delta$ is a parameter that is \emph{chosen}. The lower that $\delta$ is, the looser the state's condition is, meaning that a higher predictive confidence is needed for $(\ol{s}, \ol{a})$ to not be considered indiscriminate. In practice, we also require that we have seen $(\ol{s}, \ol{a})$ a sufficient number of times, $m$, which is also a parameter that is chosen a priori.

\begin{definition}
    Given an active feature space $\hat{F}$ and inactive feature space $\breve{F}$, a \textbf{discriminator}, $D$, is a cross product of feature sets in $\breve{F}$ that when added to the active feature space, improves the accuracy of the feedback profile by at least $\alpha \in (0,1)$, and does not cause any state that was previously not indiscriminate to become indiscriminate.
    \label{def:discriminator}
\end{definition}

These three concepts form the basis for our approach. Intuitively, our algorithm identifies indiscriminate states, finds the best possible discriminator for that indiscriminate state, and if that discriminator satisfies a set of criteria, augments the active feature space $\hat{F}$ with the discriminator. 

\subsection{Algorithm}

\noindent Algorithm~\ref{alg:main-algorithm} performs the following steps. 
\begin{enumerate}
    \item An indiscriminate state $\ol{s}$ is sampled with the additional requirement that $(\ol{s}, \ol{a})$ has been visited at least $m$ times, where $\ol{a}$ refers to the action in Definition~\ref{def:indiscriminate_state}. If no such state is found, the algorithm exits. 
    \item The full feedback dataset is split into a training set and a validation set. An element of the dataset is comprised of an action, a level of autonomy, a set of state features, and a label corresponding to the feedback received.
    \item We identify the top $k$ potential discriminator, the process for which is described below.
    \item For each discriminator $D$ found, we train a new feedback profile on the training dataset using the feature space $\hat{F} \times D$.
    \item Each feedback profile $\lambda_D$ is then tested on the validation set, and the best performing discriminator, $D^*$, is selected for validation.
    \item We validate $D^*$ by checking if $\lambda_D$ performs at least $\alpha$ better than the current $\hat{\lambda}$.
    \item Finally, if the validation is successful, the active feature space is augmented with the discriminator $D^*$, and the CAS $\mc{S}$ is updated.
\end{enumerate}

\begin{algorithm}[ht!]
    \caption{A general method for improving level-optimality in competence-aware systems}
    \label{alg:main-algorithm}
    
    \SetAlgoLined
    \DontPrintSemicolon
    \SetArgSty{textnormal}
    
    \KwIn{A dataset $\mc{D}$, a CAS $\mc{S}$, a tolerance $\delta$, and a count $k$}
    \KwResult{An updated CAS $\mc{S}$}
    \BlankLine
    $\ol{s}$ $\leftarrow$ SampleIndiscriminateState($\mc{S}, \mc{D}, \delta, \tau$)\;
    \BlankLine
    \If{$\ol{s} = \emptyset$}{Quit()\;}
    \BlankLine
    $\mc{D}_{\emph{\mbox{train}}}, \mc{D}_{\emph{\mbox{val}}} \leftarrow$ Split($\mc{D}$)\;
    \BlankLine
    \textit{discriminators} $\leftarrow$ GetDiscriminators($\mc{D}_{\emph{\mbox{train}}}$, $k$, $\ol{s}$)\;
    \For{$D$ \textbf{in} \textit{discriminators}}{
        $\lambda_D \leftarrow$ TrainClassifier($\hat{F} \times D, \mc{D}_{\emph{\mbox{train}}}$)\;
    }
    \BlankLine
    $D^* = \argmax_D$ EvaluateClassifier($\lambda_D$, $\mc{D}_{\emph{\mbox{val}}}$)\;
    \If{ValidateDiscriminator($D^*, \mc{S}$) \textbf{is} \textit{True}}{
        $\hat{F} \leftarrow \hat{F} \times D^*$\;
        Update($\mc{S}$)\;
    }
\end{algorithm}

Potential discriminators are determined as follows. First, we isolate all instances of $(\ol{s}, \ol{a})$ in the dataset $\mc{D}_{\mbox{\emph{train}}}$ and build a correlation matrix of the inactive features to the feedback that has been received for $(\ol{s}, \ol{a})$. Although different types of correlation coefficients may be used in principle, we observed that the Pearson correlation coefficient was most effective for our tasks. In particular, rank-based correlation coefficients, such as Spearman's or Kendall, did not perform well in this setting as the notion of \emph{rank} has little meaning with respect to the feedback signals or the input.

Next, we use the correlation matrix to build a discrimination matrix. The discrimination matrix provides a value, or score, for each feature set in $\breve{F}$, which is calculated by summing over the maximum absolute value for each row (feature value) in the correlation matrix associated with the feature set in question, and then normalizing over the size of the feature set. We take the absolute value because a highly negative correlation is as informative as a highly positive correlation; what matters is the magnitude of the correlation. We max over each row to prioritize having a high correlation with at least one feedback signal, rather than medium correlation with numerous feedback signals.

A final note on the algorithm is that the validation step is intended to be flexible enough to capture any requirements on adding a discriminator to the active feature space. While we only use the two stated above, there may be other factors, including domain-specific ones, that can further improve the robustness of the approach for any given domain.

\subsection{Assumptions}
\label{sec:assumptions}
In this work, we make a few key assumptions. We briefly describe these assumptions, why we made them, and why we believe they are reasonable.

First, we make the assumption that the initial transition function provided in the domain model is \emph{sufficiently correct} for any scenario where the agent is allowed, under $\kappa$, to act in an autonomous capacity. We are not concerned in this work with agents whose base domain model is poor and prone to failure, but with improving the competence of already capable systems operating in difficult domains (e.g., autonomous service robots or extraterrestrial rovers). In general, we are aiming to improve the robustness of deployed systems where the designers cannot hope to account for every possible scenario a priori, but where the scenarios that are considered are well-designed. While our approach investigates improving the competence of a CAS by augmenting the feature space to increase the granularity of the state representation, it may also be possible to increase the competence by updating the transition function itself and replanning as the human improves \emph{its} understanding of the sytem's capabilities. However, such an approach falls outside the scope of this work.

Second, we assume that the human authority has a sufficient understanding of the agent's capabilities to (1) prevent the execution of an action that the agent cannot perform successfully, and (2) provide largely consistent feedback. We make this assumption for two reasons. First, there are different ways to improve the human's understanding of the system's capabilities so that it has the appropriate trust~\cite{hoff2015trust}, or reliance, on the system. These include pre-deployment training, standardized feedback criteria, or simply expert knowledge of the system. Second, recognizing potential failure and fault recovery are separate areas of active research that is orthogonal to what we are examining in this paper. 

It is worth emphasizing that, operating under these assumptions, \emph{we do not need to update the underlying domain model's transition or reward functions directly at any point}. It suffices for the agent to simply be able to discriminate between actions that it has the competence to perform autonomously and actions that require human involvement. In instances where the agent is not competent to act autonomously, the agent will either have to do something else, or have the human take control in some capacity, and the transitions will reflect this fact. Hence the system does not need to reason about the true transition dynamics of these actions since it cannot execute them in the first place.

\subsection{Theoretical Results}
The following proposition states that if the agent and the human share the same finite set of possible features for modeling the domain, the human is consistent up to small noise in their feedback, and every state-action pair is encountered sufficiently often, then in the limit the agent will have no indiscriminate states. Intuitively, this follows from the fact that a causal feature (or more generally set of features) will always have as high of correlation as any non-causal feature, and so the algorithm will select them to be added to the model. A state can only be indiscriminate if there is some action that the feedback profile cannot confidently predict any feedback. However, if the causal features used by the human for that state-action pair are added to the model, and enough feedback is received in the limit, then eventually the feedback profile will converge sufficiently to predict the human's feedback with high probability. Hence no state will be indiscriminate in the limit. 
\begin{proposition}
    If the full feature space is finite, the human and agent share the same feature space, and the human is $\epsilon$-consistent, then if no $(\ol{s}, \ol{a}) \in \ol{S} \times \ol{A}$ is starved, in the limit for every $(\ol{s}, \ol{a}) \in \ol{S} \times \ol{A}$, there will be some $\sigma \in \Sigma$ for which it holds that $\lambda(\sigma|\ol{s},\ol{a}) > 1 - \delta$, for any $\delta \in \Big( \epsilon, 1 - \frac{1}{|\Sigma|} \Big)$.
\end{proposition}
\begin{proofsketch}
    For every $(\ol{s}, \ol{a})$, there is some subset $F^{\mc{H}}_{\ol{s},\ol{a}} \subseteq F$ that the human's feedback is dependent on for $(\ol{s}, \ol{a})$. If $F^{\mc{H}}_{\ol{s},\ol{a}} \subseteq \hat{F}$, then in the limit as no $(\ol{s}, \ol{a})$ is starved, it will be the case that $\lambda(\sigma | \ol{s}, \ol{a}) \geq 1 - \epsilon \geq 1 - \delta$, for any $\delta \in \Big( \epsilon, 1 - \frac{1}{|\Sigma|} \Big)$. So, suppose there exists some $\ol{s} \in \ol{S}$ that is indiscriminate on some action $\ol{a} \in \ol{A}$. It must be the case that some subset of $F^{\mc{H}}_{\ol{s},\ol{a}}$ is not in $\hat{F}$. However, such a subset of features, for some $\sigma \in \Sigma$, in the limit given that the human is $\epsilon$-consistent, will have correlation near $1$ and at least as high of correlation as any non-causal feature. By our algorithm, such a discriminator would be selected for evaluation and would pass the validation criteria, and so would be added to the active feature space. This is a contradiction that it is not in $\hat{F}$, so $\ol{s}$ cannot be an indiscriminate state.
\end{proofsketch}

\section{Experiments}
\label{sec:experiments}

\begin{figure}[b!]
    \centering
    \includegraphics[width=0.45\columnwidth]{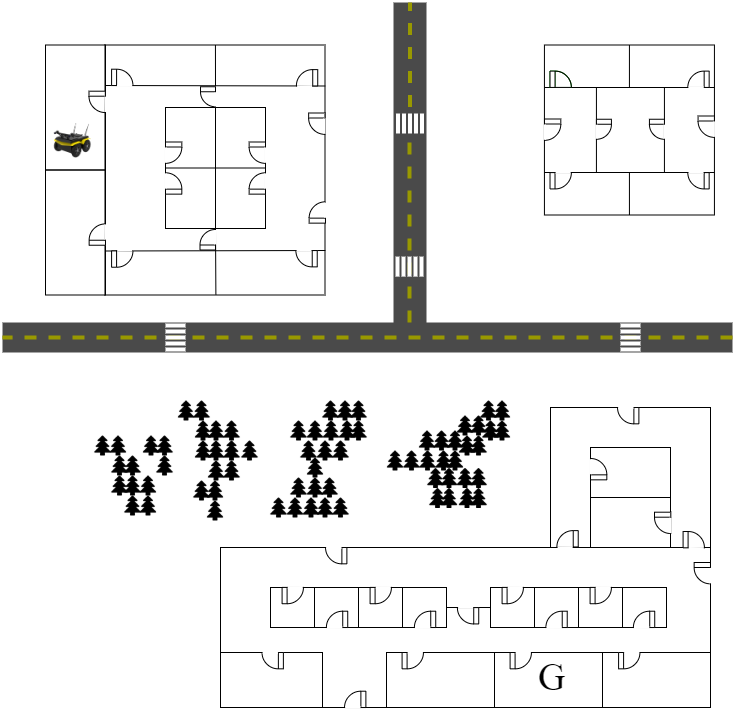}
    \caption{A depiction of the domain used for our experiments.}
    \label{fig:map}
\end{figure}

To validate our technique, we implemented our framework in a simulated domain in which a robotic agent is tasked with delivering packages to various rooms on a college campus. An illustration of the map used in our experiments can be seen in Figure~\ref{fig:map}. To accomplish its task, the agent may be required to deal with two categories of obstacles. The first is crosswalks that can have differing levels of traffic conditions that change stochastically, different visibility conditions that are fixed through time (i.e. a blind corner), and can be on either one-way streets or two-way streets, also fixed through time. The second category of obstacle is doors which the robot must open to get into buildings, through buildings, and into offices. Doors have different sizes---light, medium, and heavy---are painted various colors, and can be either push or pull doors. Trees, walls, and roads are all avoided by the system completely. Initially, $\kappa(s,a) = \{0,1\}$ for all actions in obstacle states. For states with no obstacles, the system is allowed to operate in unsupervised autonomy. In all experiments, there is a 5\% chance that the human's feedback signal is chosen uniformly at random from the possible feedback signals.

Throughout all of our experiments, we use the CAS model defined in Section~\ref{sec:example}. We consider the following four levels of autonomy: (1) no autonomy, (2) verified autonomy, (3) supervised autonomy, and (4) unsupervised autonomy. Level 1 requires a human to complete the action for the agent. Level 2 requires the agent to query the human for explicit approval prior to executing its intended action. Level 3 requires a human to be available in a supervisory capability while the agent executes the action, with the ability to override and take over control if deemed necessary. Level 4 requires no human involvement and is performed fully autonomously. The agent can receive the following four feedback signals: approval $(\oplus)$, disapproval $(\ominus)$, override $(\oslash)$, and none $(\emptyset)$. We initialize the feedback profile to a uniform prior over the feedback signals, and initialize the autonomy profile to be level 1 for any state with an obstacle in it, and level 3 otherwise.

To validate our approach, we initialize the system to only use a subset of the above features. In particular, the CAS initially only considers the traffic condition at crosswalks, and whether a door is open or closed. To increase its competence, reducing unnecessary reliance on the human, the CAS must learn which additional features it needs to add to its model to correctly predict human feedback with high confidence. In this case, the additional causal features that the human authority uses to determine their feedback are the visibility condition of the crosswalk, and both the size and opening mechanism of doors. To obfuscate things further for the system, the non-causal features are intentionally highly correlated with certain causal features. For example, in one building, the door color is perfectly consistent with the door size, and with one exception, the street type is consistent with the visibility condition. This is to test whether our approach correctly identifies causal features over correlative features.

We implement our feedback profile $\lambda$ as a GA$^2$M~\cite{lou2013accurate}. This model works well on low dimension feature spaces where pairwise feature interactions are particularly important, and has nice interpretability properties, making it desirable for robotic decision making domains. In particular, we train our $\lambda$ on all pairwise feature tensors using the native gridsearch function for hyperparameter tuning. 

In Algorithm~\ref{alg:main-algorithm}, $\delta$ is set to $0.95$, the threshold $m$ is set to 30, $k$ is set to 1, and the train/validation split is 75/25.

\subsection{Results}

In this domain, we conducted two different experiments. In the first, we set the start and goal states to be fixed throughout all episodes. In the second, the start and goal states are randomly drawn from the set of rooms through the campus map. For each experiment, we simulated both a modified CAS and a standard CAS for comparison.

\begin{figure*}[t!]
    \centering
    \begin{subfigure}[t]{0.49\columnwidth}
        \centering
        \includegraphics[width=\textwidth]{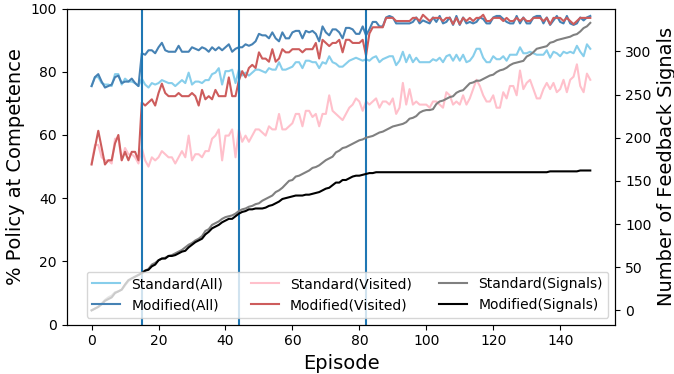}
        \caption{Start and goal states randomly assigned every episode.}
        \label{fig:random_comp}
    \end{subfigure}
    \hspace{1mm}
    \begin{subfigure}[t]{0.49\columnwidth}
        \centering
        \includegraphics[width=\textwidth]{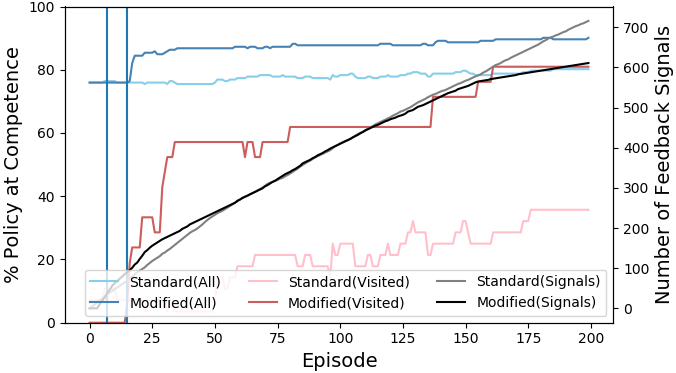}
        \caption{Start and goal states stay fixed throughout all episodes.}
        \label{fig:single_task_comp}
    \end{subfigure}
    \caption{Level-optimality of the CAS across different subsets of the state space. \emph{All states} refers to the entire state space. \emph{Visited states} are states that the system entered at least once during any episode. \emph{Cumulative signals} are the total feedback signals received after each episode. Results shown are the mean and standard error over 10 trials.}
    \label{fig:competence}
\end{figure*}

We first observe that in the modified CAS, the level-optimality reached nearly 100\% level-optimality in the random task experiment across all states, and all visited states. The vertical lines in Figure~\ref{fig:competence} indicate where discriminators were added, and we can observe that after each discriminator was added the competence starts to climb up before stagnating. While we do not expect to see this with all states in the single task, the single task experiment failed to identify and add any missing feature for crosswalks, as it only ever traversed a single crosswalk, and hence did not get enough variation in its feedback data. However, both results for the modified CAS are in stark contrast to the standard CAS which was not able to increase its level-optimality nearly as much---only a few percent across all states over the course of both experiments. In the single task domain, the standard CAS was unable to reach even 40\% level-optimality across all visited states, more than 40\% lower than the modified CAS with over 100 feedback signals more at the same point. Furthermore, the modified CAS used significantly fewer feedback signals than the standard CAS which never decreased in rate, demonstrating that the modified CAS is also less burdensome on the human. Overall, these graphs demonstrate that our method improves the CAS's competence over the course of its activity.

\begin{figure*}[t!]
    \centering
    \begin{subfigure}[t]{0.49\columnwidth}
        \centering
        \includegraphics[width=\columnwidth]{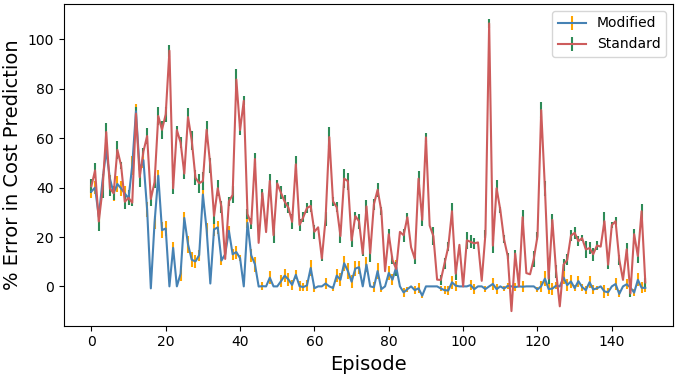}
        \caption{Start and goal states randomly assigned every episode.}
        \label{fig:random_cost}
    \end{subfigure}
    \hspace{1mm}
    \begin{subfigure}[t]{0.49\columnwidth}
        \centering
        \includegraphics[width=\columnwidth]{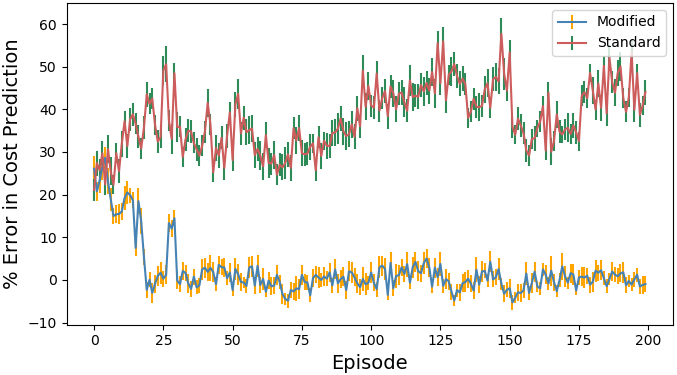}
        \caption{Start and goal states stay fixed throughout all episodes.}
        \label{fig:single_task_cost}
    \end{subfigure}
    \caption{Percent difference in the cost incurred averaged over 10 episodes, and the expected cost.}
    \label{fig:cost}
\end{figure*}

The second important result can be observed in Figure~\ref{fig:cost}. In both experiments, the expected cost converges to be almost identical to the incurred cost, averaged over 10 trials. This result demonstrates that adding the new features improves the model's accuracy and leads to better overall performance. In the standard CAS, we can see that the agent incurs costs significantly higher than predicted throughout its lifespan in both experiments, although this is most notable in the single task experiment. This indicates that the standard CAS's performance stays stagnant, while the modified CAS is able to significantly improve its performance.

\section{Related Work}
\emph{Markov Decision Processes with Uncertain Parameters}. There is a substantial body of literature on Markov decision processes (MDPs) with uncertain model parameters that has been investigated over the course of the last five decades. \emph{MDPs with imprecise transitions} (MDP-IPs) were introduced as early as 1973 by Satia and Leve~\cite{satia1973markovian}, and seminally by White and Eldeib in 1994~\cite{white1994markov}. In an MDP-IP, rather than modeling the transition dynamics as a single function, it is represented by a distribution over a set of possible transition functions. Since then, numerous expansions to the problem have been investigated including \emph{factored} MDP-IPs~\cite{delgado2011efficient} and MDPs with set-valued transitions (MDP-STs)~\cite{trevizan2007planning}, as well as novel solution approaches, including a multi-linear and integer programming approach~\cite{shirota2007multilinear}, and a real-time dynamic programming approach~\cite{delgado2016real}.

Similarly, MDPs with uncertain rewards (sometimes referred to as IR-MDPs, or \emph{imprecise reward MDPs}) were introduced as early as 1986 by White and Eldeib~\cite{white1986parameter}, and are similar in concept to the MDP-IP. More recently, Regan and Boutilier have proposed both an offline~\cite{regan2010robust} and online~\cite{regan2011robust} approach to solving IR-MDPs, which are, to our knowledge, the state of the art in computational solutions to this class of problems.

However, more generally, the concept of an \emph{uncertain MDP} (UMDP) was first formally described by Chen and Bowling~\cite{chen2012tractable} to capture MDPs in which both the transitions and rewards may be uncertain or imprecisely specified. Note that this is not to be confused with MDP-IPs referred to as uMDPs~\cite{bagnell2001solving, wolff2012robust} or IR-MDPs referred to as uMDPs~\cite{xu2009parametric}), and has since been expanded upon~\cite{ahmed2013regret, adulyasak2015solving, ahmed2017sampling}.

All of these models are intended to deal with domains where producing a fully precise and accurate model of the domain a priori is hard or infeasible, or where the dynamics are simply non-stationary. While the motivation of these lines of work are similar to ours, namely the fact that precisely modeling complex, real-world domains is often hard or impossible to do prior to deployment, our work differs starkly in that we do not directly modify the transition or reward functions, but rather we are concerned with modifying the factored state representation to better match that of a separate entity (e.g. the human). 
\\\\
\emph{Learning from Human Interaction}. There are several areas of research that have investigated learning from interaction with a human, namely \emph{learning from human input}~\cite{rosenstein2004supervised, suay2011effect, torrey2013teaching} and \emph{learning from demonstration}~\cite{chernova2009interactive, clouse1997integrating, del2018not, rigter2020framework}. In the former, the agent must learn some parameter of the model, generally the q-value of an action or the best action itself at a given state, from interacting with a human either through advice or questions. In the latter, the agent attempts to learn its policy by mimicking the actions of another agent (generally a human).

However, while our work is concerned with learning from interacting with a human, both the focus and scope of what we are attempting to have the agent learn is very different as we are limiting the learning to be only over missing features \emph{used by the human in their feedback}.
\\\\
\emph{Model-Based Reinforcement Learning}. Our problem is related to the work on model-based RL in which the agent actively builds an estimate of the transition and reward function for the domain it is operating in without relying on a model to be defined a priori~\cite{williams2017information, walsh2010integrating}. By bypassing the need for an a priori model entirely, model-based RL as a decision-making framework does not suffer from the primary problem investigated in this paper, i.e. imprecise and incomplete models. In fact, there is large body of literature that has investigated ways to use reinforcement learning for robot decision making that addresses some of its primary challenges such as safety and explainability~\cite{anjomshoae2019explainable,berkenkamp2017safe,smart2002effective}. However, our work is geared towards improving robotic systems which use stochastic model-based planning methods, rather than reinforcement learning. 
\\\\
\emph{Feature Selection}. Given that a primary element of our work is identifying useful features, by some metric, from some large set of possible features, our work is highly related to that of unsupervised feature selection where the objective is to select from large space of possible features the one(s) that maximize some given objective function~\cite{dash1997feature}. Feature selection has most commonly been applied to data analysis where the data is high dimensional, warranting a need to select the features that optimize the objective function while preserving the underlying structure of the data~\cite{du2015unsupervised, nie2016unsupervised}. It has also been used in reinforcement learning as a technique to more efficiently determine the most relevant sensory features for capturing the transition dynamics of the domain~\cite{diuk2009adaptive}. Largely, because the \emph{method} of identifying the features is not the focus of our work, feature selection is an orthogonal problem to what is studied here. However, we believe that many ideas from this area are symbiotic with our work, and we will be examining how we may incorporate these methods to improve the performance of our approach.

\section{Discussion}

This paper proposes a method for providing competence-aware systems the ability to improve their competence over time by identifying key features missing from their model of the domain and adding those features into their state representation. This method works by identifying \emph{indiscriminate states}, states missing information needed to predict human feedback, and for such states determining the most likely \emph{discriminators}, features that improve the confidence of the feedback profile by enabling the system to better discriminate the feedback for indiscriminate states. To validate our approach, we provide empirical results in a simulated delivery robot domain that demonstrate that our approach (1) identifies the missing features, (2) only adds the missing \emph{causal} features, (3) significantly improves the competence of the system when compared to an unmodified CAS, and (4) leads to major improvements in performance. However, there are several areas of potential improvement for our approach that we are actively addressing for future work.
\\\\
\emph{Human Understanding of System Ability}. Our method relies on the assumption that the human has a sufficient understanding of the system's underlying capabilities to consistently provide accurate feedback, and to know when to disallow the agent to perform certain actions autonomously that it is not capable of doing successfully. In practice, these assumptions may not be the case. To address this issue, we are investigating the use of new feedback signals that specifically indicate a lack of knowledge or confidence on the part of the human, prompting the system to gather additional information rather than to act immediately. This is similar to the recently proposed approach where an agent must monitor its own ability and consequently determine where to ask for help from a human (which has a cost), and learn from the provided demonstration~\cite{rigter2020framework}. 
\\\\
\emph{Feedback from Multiple Human Authorities.} Similarly, we make the assumption that feedback comes from a single source---namely, a single human authority. In practice it may often be the case that the system interacts with, and receives feedback from, different human authorities each with different perspectives, preferences, or simply different understanding of the system's abilities. In practice, this problem can be addressed in some cases by requiring a training phase for any authority that interacts with the system to provide a rigorous understanding of the system's capabilities, and a set of standardized feedback criteria to enforce consistency of feedback. Hence, we could extend the CAS model to not have a single feedback profile $\lambda$, but instead to maintain a distribution over feedback profiles, or learn a different profile for each human that interacts with the system to better personalize the response.
\\\\
\emph{Incongruous Domain Models}. In general, it may not be the case that every, or even any, feature that the human authority's feedback is conditioned on
%%, i.e. every causal feature, 
is available to the agent. In these cases, it may still be beneficial to identify and add correlative features to the model as proxies for the causal features. However, these features may cause problems elsewhere in the domain where the correlation does hot hold, so addressing this issue is key in making this method more robust.
\\\\
\emph{Practical Models}. The preliminary results presented in this paper are on a simple domain with only a small number of (highly abstract) features. We are testing our approach on a real mobile robot where our method must operate directly on the high dimensional \emph{sensory features} available to the agent, rather than semantically interpretable features as presented in the paper. By replacing the GA$^2$M used for the feedback profile in this work with a more sophisticated CNN, we may be able to identify these discriminators directly from the perceptual data.
\\\\
\emph{Feature Selection}. As noted earlier, there is a large body of literature on the problem of feature selection which our approach largely does not utilize as presented here. We will investigate ways to incorporate techniques from feature selection to improve both the efficiency and effectiveness of our algorithm with respect to identifying missing causal features. This problem is particularly relevant in high dimensional feature spaces like the sensory features of a mobile robot, where the computational overhead will be much higher, and there is more noise in the data.
%\\\\
%\emph{Multi-Agent Systems}. Finally, we are interested in extending this framework to the multi-agent system in two ways. First, in large scale multi-agent systems, such as a group of semi-autonomous robots, it is likely that different human-agent systems will have different competence, driven either by experience or even preference. However, while the competence of two different systems may differ, it is also likely that feedback provided in one system is useful in others. To incorporate this, we are experimenting with a notion of \emph{soft feedback}, in which feedback can be shared across multiple CAS in one multi-agent system, weighted by a similarity measure between the learned competence of the systems.

\section*{Acknowledgments}
This work was supported in part by the National Science Foundation Grants IIS-1724101, IIS-1813490, and DGE-1451512, and in part by the Alliance Innovation Lab Silicon Valley.

\bibliographystyle{eptcs}

\begin{thebibliography}{10}
	\providecommand{\bibitemdeclare}[2]{}
	\providecommand{\surnamestart}{}
	\providecommand{\surnameend}{}
	\providecommand{\urlprefix}{Available at }
	\providecommand{\url}[1]{\texttt{#1}}
	\providecommand{\href}[2]{\texttt{#2}}
	\providecommand{\urlalt}[2]{\href{#1}{#2}}
	\providecommand{\doi}[1]{doi:\urlalt{http://dx.doi.org/#1}{#1}}
	\providecommand{\bibinfo}[2]{#2}
	
	\bibitemdeclare{inproceedings}{adulyasak2015solving}
	\bibitem{adulyasak2015solving}
	\bibinfo{author}{Yossiri \surnamestart Adulyasak\surnameend},
	\bibinfo{author}{Pradeep \surnamestart Varakantham\surnameend},
	\bibinfo{author}{Asrar \surnamestart Ahmed\surnameend} \&
	\bibinfo{author}{Patrick \surnamestart Jaillet\surnameend}
	(\bibinfo{year}{2015}): \emph{\bibinfo{title}{Solving uncertain MDPs with
			objectives that are separable over instantiations of model uncertainty}}.
	\newblock In: {\sl \bibinfo{booktitle}{AAAI Conference on Artificial
			Intelligence}}, pp. \bibinfo{pages}{3454--3460}.
	
	\bibitemdeclare{inproceedings}{ahmed2013regret}
	\bibitem{ahmed2013regret}
	\bibinfo{author}{Asrar \surnamestart Ahmed\surnameend},
	\bibinfo{author}{Pradeep \surnamestart Varakantham\surnameend},
	\bibinfo{author}{Yossiri \surnamestart Adulyasak\surnameend} \&
	\bibinfo{author}{Patrick \surnamestart Jaillet\surnameend}
	(\bibinfo{year}{2013}): \emph{\bibinfo{title}{Regret based robust solutions
			for uncertain Markov decision processes}}.
	\newblock In: {\sl \bibinfo{booktitle}{Advances in Neural Information
			Processing Systems (NeurIPS)}}, pp. \bibinfo{pages}{881--889},
	\doi{10.1613/jai.5242}.
	
	\bibitemdeclare{article}{ahmed2017sampling}
	\bibitem{ahmed2017sampling}
	\bibinfo{author}{Asrar \surnamestart Ahmed\surnameend},
	\bibinfo{author}{Pradeep \surnamestart Varakantham\surnameend},
	\bibinfo{author}{Meghna \surnamestart Lowalekar\surnameend},
	\bibinfo{author}{Yossiri \surnamestart Adulyasak\surnameend} \&
	\bibinfo{author}{Patrick \surnamestart Jaillet\surnameend}
	(\bibinfo{year}{2017}): \emph{\bibinfo{title}{Sampling based approaches for
			minimizing regret in uncertain Markov decision processes (MDPs)}}.
	\newblock {\sl \bibinfo{journal}{Journal of Artificial Intelligence Research
			(JAIR)}} \bibinfo{volume}{59}, pp. \bibinfo{pages}{229--264},
	\doi{10.1613/jair.5242}.
	
	\bibitemdeclare{inproceedings}{anjomshoae2019explainable}
	\bibitem{anjomshoae2019explainable}
	\bibinfo{author}{Sule \surnamestart Anjomshoae\surnameend},
	\bibinfo{author}{Amro \surnamestart Najjar\surnameend},
	\bibinfo{author}{Davide \surnamestart Calvaresi\surnameend} \&
	\bibinfo{author}{Kary \surnamestart Fr{\"a}mling\surnameend}
	(\bibinfo{year}{2019}): \emph{\bibinfo{title}{Explainable agents and robots:
			Results from a systematic literature review}}.
	\newblock In: {\sl \bibinfo{booktitle}{International Conference on Autonomous
			Agents and MultiAgent Systems (AAMAS)}}, pp. \bibinfo{pages}{1078--1088}.
	
	\bibitemdeclare{techreport}{bagnell2001solving}
	\bibitem{bagnell2001solving}
	\bibinfo{author}{J.~Andrew \surnamestart Bagnell\surnameend},
	\bibinfo{author}{Andrew~Y. \surnamestart Ng\surnameend} \&
	\bibinfo{author}{Jeff~G. \surnamestart Schneider\surnameend}
	(\bibinfo{year}{2001}): \emph{\bibinfo{title}{Solving uncertain Markov
			decision processes}}.
	\newblock \bibinfo{type}{Technical Report}, \bibinfo{institution}{Carnegie
		Mellon University}.
	
	\bibitemdeclare{inproceedings}{basich2020learning}
	\bibitem{basich2020learning}
	\bibinfo{author}{Connor \surnamestart Basich\surnameend},
	\bibinfo{author}{Justin \surnamestart Svegliato\surnameend},
	\bibinfo{author}{Kyle~Hollins \surnamestart Wray\surnameend},
	\bibinfo{author}{Stefan \surnamestart Witwicki\surnameend},
	\bibinfo{author}{Joydeep \surnamestart Biswas\surnameend} \&
	\bibinfo{author}{Shlomo \surnamestart Zilberstein\surnameend}
	(\bibinfo{year}{2020}): \emph{\bibinfo{title}{Learning to Optimize Autonomy
			in Competence-Aware Systems}}.
	\newblock In: {\sl \bibinfo{booktitle}{International Joint Conference on
			Autonomous Agents and Multiagent Systems (AAMAS)}}, pp.
	\bibinfo{pages}{123--131}.
	
	\bibitemdeclare{inproceedings}{berkenkamp2017safe}
	\bibitem{berkenkamp2017safe}
	\bibinfo{author}{Felix \surnamestart Berkenkamp\surnameend},
	\bibinfo{author}{Matteo \surnamestart Turchetta\surnameend},
	\bibinfo{author}{Angela \surnamestart Schoellig\surnameend} \&
	\bibinfo{author}{Andreas \surnamestart Krause\surnameend}
	(\bibinfo{year}{2017}): \emph{\bibinfo{title}{Safe model-based reinforcement
			learning with stability guarantees}}.
	\newblock In: {\sl \bibinfo{booktitle}{Advances in Neural Information
			Processing Systems (NeurIPS)}}, pp. \bibinfo{pages}{908--918}.
	
	\bibitemdeclare{article}{broggi1999argo}
	\bibitem{broggi1999argo}
	\bibinfo{author}{Alberto \surnamestart Broggi\surnameend},
	\bibinfo{author}{Massimo \surnamestart Bertozzi\surnameend},
	\bibinfo{author}{Alessandra \surnamestart Fascioli\surnameend},
	\bibinfo{author}{C.~Guarino~Lo \surnamestart Bianco\surnameend} \&
	\bibinfo{author}{Aurelio \surnamestart Piazzi\surnameend}
	(\bibinfo{year}{1999}): \emph{\bibinfo{title}{The {ARGO} autonomous
			vehicle’s vision and control systems}}.
	\newblock {\sl \bibinfo{journal}{International Journal of Intelligent Control
			and Systems}} \bibinfo{volume}{3}(\bibinfo{number}{4}), pp.
	\bibinfo{pages}{409--441}.
	
	\bibitemdeclare{article}{broggi2012vislab}
	\bibitem{broggi2012vislab}
	\bibinfo{author}{Alberto \surnamestart Broggi\surnameend},
	\bibinfo{author}{Pietro \surnamestart Cerri\surnameend},
	\bibinfo{author}{Mirko \surnamestart Felisa\surnameend},
	\bibinfo{author}{Maria~Chiara \surnamestart Laghi\surnameend},
	\bibinfo{author}{Luca \surnamestart Mazzei\surnameend} \&
	\bibinfo{author}{Pier~Paolo \surnamestart Porta\surnameend}
	(\bibinfo{year}{2012}): \emph{\bibinfo{title}{The {VisLab} Intercontinental
			Autonomous Challenge: An extensive test for a platoon of intelligent
			vehicles}}.
	\newblock {\sl \bibinfo{journal}{International Journal of Vehicle Autonomous
			Systems}} \bibinfo{volume}{10}(\bibinfo{number}{3}), pp.
	\bibinfo{pages}{147--164}, \doi{10.1504/IJVAS.2012.051250}.
	
	\bibitemdeclare{inproceedings}{chen2012tractable}
	\bibitem{chen2012tractable}
	\bibinfo{author}{Katherine \surnamestart Chen\surnameend} \&
	\bibinfo{author}{Michael \surnamestart Bowling\surnameend}
	(\bibinfo{year}{2012}): \emph{\bibinfo{title}{Tractable objectives for robust
			policy optimization}}.
	\newblock In: {\sl \bibinfo{booktitle}{Advances in Neural Information
			Processing Systems (NeurIPS)}}, pp. \bibinfo{pages}{2069--2077}.
	
	\bibitemdeclare{article}{chernova2009interactive}
	\bibitem{chernova2009interactive}
	\bibinfo{author}{Sonia \surnamestart Chernova\surnameend} \&
	\bibinfo{author}{Manuela \surnamestart Veloso\surnameend}
	(\bibinfo{year}{2009}): \emph{\bibinfo{title}{Interactive policy learning
			through confidence-based autonomy}}.
	\newblock {\sl \bibinfo{journal}{Journal of Artificial Intelligence Research
			(JAIR)}} \bibinfo{volume}{34}, pp. \bibinfo{pages}{1--25},
	\doi{10.1613/jair.2584}.
	
	\bibitemdeclare{phdthesis}{clouse1997integrating}
	\bibitem{clouse1997integrating}
	\bibinfo{author}{Jeffery~A. \surnamestart Clouse\surnameend}
	(\bibinfo{year}{1996}): \emph{\bibinfo{title}{On integrating apprentice
			learning and reinforcement learning}}.
	\newblock Ph.D. thesis, \bibinfo{school}{University of Massachusetts Amherst}.
	
	\bibitemdeclare{article}{dash1997feature}
	\bibitem{dash1997feature}
	\bibinfo{author}{Manoranjan \surnamestart Dash\surnameend} \&
	\bibinfo{author}{Huan \surnamestart Liu\surnameend} (\bibinfo{year}{1997}):
	\emph{\bibinfo{title}{Feature selection for classification}}.
	\newblock {\sl \bibinfo{journal}{Intelligent Data Analysis}}
	\bibinfo{volume}{1}(\bibinfo{number}{3}), pp. \bibinfo{pages}{131--156},
	\doi{10.1016/S1088-467X(97)00008-5}.
	
	\bibitemdeclare{article}{del2018not}
	\bibitem{del2018not}
	\bibinfo{author}{Francesco \surnamestart Del~Duchetto\surnameend},
	\bibinfo{author}{Ayse \surnamestart Kucukyilmaz\surnameend},
	\bibinfo{author}{Luca \surnamestart Iocchi\surnameend} \&
	\bibinfo{author}{Marc \surnamestart Hanheide\surnameend}
	(\bibinfo{year}{2018}): \emph{\bibinfo{title}{Do not make the same mistakes
			again and again: Learning local recovery policies for navigation from human
			demonstrations}}.
	\newblock {\sl \bibinfo{journal}{IEEE Robotics and Automation Letters (RA-L)}}
	\bibinfo{volume}{3}(\bibinfo{number}{4}), pp. \bibinfo{pages}{4084--4091},
	\doi{10.1109/LRA.2018.2861080}
	
	\bibitemdeclare{article}{delgado2016real}
	\bibitem{delgado2016real}
	\bibinfo{author}{Karina~V. \surnamestart Delgado\surnameend},
	\bibinfo{author}{Leliane~N. \surnamestart De~Barros\surnameend},
	\bibinfo{author}{Daniel~B. \surnamestart Dias\surnameend} \&
	\bibinfo{author}{Scott \surnamestart Sanner\surnameend}
	(\bibinfo{year}{2016}): \emph{\bibinfo{title}{Real-time dynamic programming
			for Markov decision processes with imprecise probabilities}}.
	\newblock {\sl \bibinfo{journal}{Artificial Intelligence (AIJ)}}
	\bibinfo{volume}{230}, pp. \bibinfo{pages}{192--223},
	\doi{10.1016/j.artint.2015.09.005}.
	
	\bibitemdeclare{article}{delgado2011efficient}
	\bibitem{delgado2011efficient}
	\bibinfo{author}{Karina~Valdivia \surnamestart Delgado\surnameend},
	\bibinfo{author}{Scott \surnamestart Sanner\surnameend} \&
	\bibinfo{author}{Leliane~Nunes \surnamestart De~Barros\surnameend}
	(\bibinfo{year}{2011}): \emph{\bibinfo{title}{Efficient solutions to factored
			MDPs with imprecise transition probabilities}}.
	\newblock {\sl \bibinfo{journal}{Artificial Intelligence (AIJ)}}
	\bibinfo{volume}{175}(\bibinfo{number}{9-10}), pp.
	\bibinfo{pages}{1498--1527}, \doi{10.1016/j.artint.2011.01.001}.
	
	\bibitemdeclare{book}{dickmanns2007dynamic}
	\bibitem{dickmanns2007dynamic}
	\bibinfo{author}{Ernst~D. \surnamestart Dickmanns\surnameend}
	(\bibinfo{year}{2007}): \emph{\bibinfo{title}{Dynamic vision for perception
			and control of motion}}.
	\newblock \bibinfo{publisher}{Springer Science \& Business Media},
	\doi{10.1007/978-1-84628-638-4}.
	
	\bibitemdeclare{inproceedings}{diuk2009adaptive}
	\bibitem{diuk2009adaptive}
	\bibinfo{author}{Carlos \surnamestart Diuk\surnameend}, \bibinfo{author}{Lihong
		\surnamestart Li\surnameend} \& \bibinfo{author}{Bethany~R. \surnamestart
		Leffler\surnameend} (\bibinfo{year}{2009}): \emph{\bibinfo{title}{The
			adaptive k-meteorologists problem and its application to structure learning
			and feature selection in reinforcement learning}}.
	\newblock In: {\sl \bibinfo{booktitle}{International Conference on Machine
			Learning (ICML)}}, pp. \bibinfo{pages}{249--256},
	\doi{10.1145/1553374.1553406}.
	
	\bibitemdeclare{inproceedings}{du2015unsupervised}
	\bibitem{du2015unsupervised}
	\bibinfo{author}{Liang \surnamestart Du\surnameend} \& \bibinfo{author}{Yi-Dong
		\surnamestart Shen\surnameend} (\bibinfo{year}{2015}):
	\emph{\bibinfo{title}{Unsupervised feature selection with adaptive structure
			learning}}.
	\newblock In: {\sl \bibinfo{booktitle}{ACM International Conference on
			Knowledge Discovery and Data Mining (SIGKDD)}}, pp.
	\bibinfo{pages}{209--218}, \doi{10.1145/2783258.2783345}.
	
	\bibitemdeclare{article}{gao2017review}
	\bibitem{gao2017review}
	\bibinfo{author}{Yang \surnamestart Gao\surnameend} \& \bibinfo{author}{Steve
		\surnamestart Chien\surnameend} (\bibinfo{year}{2017}):
	\emph{\bibinfo{title}{Review on space robotics: Toward top-level science
			through space exploration}}.
	\newblock {\sl \bibinfo{journal}{Science Robotics}}
	\bibinfo{volume}{2}(\bibinfo{number}{7}), \doi{10.1126/scirobotics.aan5074}.
	
	\bibitemdeclare{article}{hawes2017strands}
	\bibitem{hawes2017strands}
	\bibinfo{author}{Nick \surnamestart Hawes\surnameend},
	\bibinfo{author}{Christopher \surnamestart Burbridge\surnameend},
	\bibinfo{author}{Ferdian \surnamestart Jovan\surnameend},
	\bibinfo{author}{Lars \surnamestart Kunze\surnameend}, \bibinfo{author}{Bruno
		\surnamestart Lacerda\surnameend}, \bibinfo{author}{Lenka \surnamestart
		Mudrova\surnameend}, \bibinfo{author}{Jay \surnamestart Young\surnameend},
	\bibinfo{author}{Jeremy \surnamestart Wyatt\surnameend},
	\bibinfo{author}{Denise \surnamestart Hebesberger\surnameend},
	\bibinfo{author}{Tobias \surnamestart Kortner\surnameend} et~al.
	(\bibinfo{year}{2017}): \emph{\bibinfo{title}{The {STRANDS} project:
			Long-term autonomy in everyday environments}}.
	\newblock {\sl \bibinfo{journal}{IEEE Robotics \& Automation Magazine}}
	\bibinfo{volume}{24}(\bibinfo{number}{3}), \doi{10.1109/MRA.2016.2636359}.
	
	\bibitemdeclare{article}{hoff2015trust}
	\bibitem{hoff2015trust}
	\bibinfo{author}{Kevin~Anthony \surnamestart Hoff\surnameend} \&
	\bibinfo{author}{Masooda \surnamestart Bashir\surnameend}
	(\bibinfo{year}{2015}): \emph{\bibinfo{title}{Trust in automation:
			Integrating empirical evidence on factors that influence trust}}.
	\newblock {\sl \bibinfo{journal}{Human Factors}}
	\bibinfo{volume}{57}(\bibinfo{number}{3}), pp. \bibinfo{pages}{407--434},
	\doi{10.1177/0018720814547570}.
	
	\bibitemdeclare{inproceedings}{lou2013accurate}
	\bibitem{lou2013accurate}
	\bibinfo{author}{Yin \surnamestart Lou\surnameend}, \bibinfo{author}{Rich
		\surnamestart Caruana\surnameend}, \bibinfo{author}{Johannes \surnamestart
		Gehrke\surnameend} \& \bibinfo{author}{Giles \surnamestart Hooker\surnameend}
	(\bibinfo{year}{2013}): \emph{\bibinfo{title}{Accurate intelligible models
			with pairwise interactions}}.
	\newblock In: {\sl \bibinfo{booktitle}{ACM International Conference on
			Knowledge Discovery and Data Mining (SIGKDD)}}, pp.
	\bibinfo{pages}{623--631}, \doi{10.1145/2487575.2487579}.
	
	\bibitemdeclare{inproceedings}{meeussen2011long}
	\bibitem{meeussen2011long}
	\bibinfo{author}{Wim \surnamestart Meeussen\surnameend}, \bibinfo{author}{Eitan
		\surnamestart Marder-Eppstein\surnameend}, \bibinfo{author}{Kevin
		\surnamestart Watts\surnameend} \& \bibinfo{author}{Brian~P. \surnamestart
		Gerkey\surnameend} (\bibinfo{year}{2011}): \emph{\bibinfo{title}{Long term
			autonomy in office environments}}.
	\newblock In: {\sl \bibinfo{booktitle}{Robotics: Science and Systems (RSS)
			{ALONE} Workshop}}.
	
	\bibitemdeclare{inproceedings}{mustard2013mars}
	\bibitem{mustard2013mars}
	\bibinfo{author}{John~F. \surnamestart Mustard\surnameend},
	\bibinfo{author}{D.~\surnamestart Beaty\surnameend} \&
	\bibinfo{author}{D.~\surnamestart Bass\surnameend} (\bibinfo{year}{2013}):
	\emph{\bibinfo{title}{Mars 2020 science rover: Science goals and mission
			concept}}.
	\newblock In: {\sl \bibinfo{booktitle}{AAS/Division for Planetary Sciences
			Meeting Abstracts}}, \bibinfo{volume}{45}.
	
	\bibitemdeclare{inproceedings}{nie2016unsupervised}
	\bibitem{nie2016unsupervised}
	\bibinfo{author}{Feiping \surnamestart Nie\surnameend}, \bibinfo{author}{Wei
		\surnamestart Zhu\surnameend} \& \bibinfo{author}{Xuelong \surnamestart
		Li\surnameend} (\bibinfo{year}{2016}): \emph{\bibinfo{title}{Unsupervised
			feature selection with structured graph optimization}}.
	\newblock In: {\sl \bibinfo{booktitle}{AAAI Conference on Artificial
			Intelligence}}, pp. \bibinfo{pages}{1302--1308}.
	
	\bibitemdeclare{inproceedings}{pineda2015continual}
	\bibitem{pineda2015continual}
	\bibinfo{author}{Luis \surnamestart Pineda\surnameend},
	\bibinfo{author}{Takeshi \surnamestart Takahashi\surnameend},
	\bibinfo{author}{Hee-Tae \surnamestart Jung\surnameend},
	\bibinfo{author}{Shlomo \surnamestart Zilberstein\surnameend} \&
	\bibinfo{author}{Roderic \surnamestart Grupen\surnameend}
	(\bibinfo{year}{2015}): \emph{\bibinfo{title}{Continual planning for search
			and rescue robots}}.
	\newblock In: {\sl \bibinfo{booktitle}{IEEE/RAS International Conference on
			Humanoid Robots (Humanoids)}}, \bibinfo{organization}{IEEE}, pp.
	\bibinfo{pages}{243--248}, \doi{10.1109/HUMANOIDS.2015.7363542}.
	
	\bibitemdeclare{inproceedings}{regan2010robust}
	\bibitem{regan2010robust}
	\bibinfo{author}{Kevin \surnamestart Regan\surnameend} \&
	\bibinfo{author}{Craig \surnamestart Boutilier\surnameend}
	(\bibinfo{year}{2010}): \emph{\bibinfo{title}{Robust policy computation in
			reward-uncertain MDPs using nondominated policies}}.
	\newblock In: {\sl \bibinfo{booktitle}{AAAI Conference on Artificial
			Intelligence}}, pp. \bibinfo{pages}{1127--1133}.
	
	\bibitemdeclare{inproceedings}{regan2011robust}
	\bibitem{regan2011robust}
	\bibinfo{author}{Kevin \surnamestart Regan\surnameend} \&
	\bibinfo{author}{Craig \surnamestart Boutilier\surnameend}
	(\bibinfo{year}{2011}): \emph{\bibinfo{title}{Robust online optimization of
			reward-uncertain MDPs}}.
	\newblock In: {\sl \bibinfo{booktitle}{International Joint Conference on
			Artificial Intelligence (IJCAI)}}, pp. \bibinfo{pages}{2165--2171},
	\doi{10.5591/978-1-57735-516-8/IJCAI11-361}.
	
	\bibitemdeclare{article}{rigter2020framework}
	\bibitem{rigter2020framework}
	\bibinfo{author}{Marc \surnamestart Rigter\surnameend}, \bibinfo{author}{Bruno
		\surnamestart Lacerda\surnameend} \& \bibinfo{author}{Nick \surnamestart
		Hawes\surnameend} (\bibinfo{year}{2020}): \emph{\bibinfo{title}{A Framework
			for learning From demonstration with minimal human effort}}.
	\newblock {\sl \bibinfo{journal}{IEEE Robotics and Automation Letters (RA-L)}}
	\bibinfo{volume}{5}(\bibinfo{number}{2}), pp. \bibinfo{pages}{2023--2030},
	\doi{10.1109/LRA.2018.2861080}.
	
	\bibitemdeclare{incollection}{rosenstein2004supervised}
	\bibitem{rosenstein2004supervised}
	\bibinfo{author}{Michael~T. \surnamestart Rosenstein\surnameend} \&
	\bibinfo{author}{Andrew~G. \surnamestart Barto\surnameend}
	(\bibinfo{year}{2004}): \emph{\bibinfo{title}{Supervised actor-critic
			reinforcement learning}}.
	\newblock In \bibinfo{editor}{J.~\surnamestart Si\surnameend},
	\bibinfo{editor}{A.~G. \surnamestart Barto\surnameend},
	\bibinfo{editor}{W.~B. \surnamestart Powell\surnameend} \&
	\bibinfo{editor}{D.~\surnamestart Wunsch\surnameend}, editors: {\sl
		\bibinfo{booktitle}{Handbook of Learning and Approximate Dynamic
			Programming}}, chapter~\bibinfo{chapter}{7}, \bibinfo{publisher}{IEEE Press},
	pp. \bibinfo{pages}{359--380}.
	
	\bibitemdeclare{inproceedings}{saisubramanian2017optimizing}
	\bibitem{saisubramanian2017optimizing}
	\bibinfo{author}{Sandhya \surnamestart Saisubramanian\surnameend},
	\bibinfo{author}{Shlomo \surnamestart Zilberstein\surnameend} \&
	\bibinfo{author}{Prashant \surnamestart Shenoy\surnameend}
	(\bibinfo{year}{2017}): \emph{\bibinfo{title}{Optimizing electric vehicle
			charging through determinization}}.
	\newblock In: {\sl \bibinfo{booktitle}{ICAPS Workshop on Scheduling and
			Planning Applications}}.
	
	\bibitemdeclare{article}{satia1973markovian}
	\bibitem{satia1973markovian}
	\bibinfo{author}{Jay~K. \surnamestart Satia\surnameend} \&
	\bibinfo{author}{Roy~E. \surnamestart Lave~Jr.\surnameend}
	(\bibinfo{year}{1973}): \emph{\bibinfo{title}{Markovian decision processes
			with uncertain transition probabilities}}.
	\newblock {\sl \bibinfo{journal}{Operations Research}}
	\bibinfo{volume}{21}(\bibinfo{number}{3}), pp. \bibinfo{pages}{728--740},
	\doi{10.1287/opre.21.3.728}.
	
	\bibitemdeclare{inproceedings}{shirota2007multilinear}
	\bibitem{shirota2007multilinear}
	\bibinfo{author}{Ricardo \surnamestart Shirota~Filho\surnameend},
	\bibinfo{author}{Fabio~Gagliardi \surnamestart Cozman\surnameend},
	\bibinfo{author}{Felipe~W. \surnamestart Trevizan\surnameend},
	\bibinfo{author}{Cassio~Polpo \surnamestart de~Campos\surnameend} \&
	\bibinfo{author}{Leliane~Nunes \surnamestart De~Barros\surnameend}
	(\bibinfo{year}{2007}): \emph{\bibinfo{title}{Multilinear and integer
			programming for Markov decision processes with imprecise probabilities}}.
	\newblock In: {\sl \bibinfo{booktitle}{5th International Symposium on Imprecise Probability: Theories and Applications}} pp. \bibinfo{pages}{395--404}.
	
	\bibitemdeclare{inproceedings}{smart2002effective}
	\bibitem{smart2002effective}
	\bibinfo{author}{William~D. \surnamestart Smart\surnameend} \&
	\bibinfo{author}{Leslie~Pack \surnamestart Kaelbling\surnameend}
	(\bibinfo{year}{2002}): \emph{\bibinfo{title}{Effective reinforcement
			learning for mobile robots}}.
	\newblock In: {\sl \bibinfo{booktitle}{IEEE International Conference on
			Robotics and Automation (ICRA)}}, \bibinfo{volume}{4}, pp.
	\bibinfo{pages}{3404--3410}, \doi{10.1109/ROBOT.2002.1014237}.
	
	\bibitemdeclare{inproceedings}{suay2011effect}
	\bibitem{suay2011effect}
	\bibinfo{author}{Halit~Bener \surnamestart Suay\surnameend} \&
	\bibinfo{author}{Sonia \surnamestart Chernova\surnameend}
	(\bibinfo{year}{2011}): \emph{\bibinfo{title}{Effect of human guidance and
			state space size on interactive reinforcement learning}}.
	\newblock In: {\sl \bibinfo{booktitle}{IEEE Conference on Robot and Human
			Interactive Communication (RO-MAN)}}, \bibinfo{organization}{IEEE}, pp.
	\bibinfo{pages}{1--6}, \doi{10.1109/ROMAN.2011.6005223}.
	
	\bibitemdeclare{inproceedings}{svegliato2019belief}
	\bibitem{svegliato2019belief}
	\bibinfo{author}{Justin \surnamestart Svegliato\surnameend},
	\bibinfo{author}{Kyle~Hollins \surnamestart Wray\surnameend},
	\bibinfo{author}{Stefan~J. \surnamestart Witwicki\surnameend},
	\bibinfo{author}{Joydeep \surnamestart Biswas\surnameend} \&
	\bibinfo{author}{Shlomo \surnamestart Zilberstein\surnameend}
	(\bibinfo{year}{2019}): \emph{\bibinfo{title}{Belief Space Metareasoning for
			Exception Recovery}}.
	\newblock In: {\sl \bibinfo{booktitle}{IEEE/RSJ International Conference on
			Intelligent Robots and Systems (IROS)}}, pp. \bibinfo{pages}{1224--1229},
	\doi{10.1109/IROS40897.2019.8967676}.
	
	\bibitemdeclare{inproceedings}{torrey2013teaching}
	\bibitem{torrey2013teaching}
	\bibinfo{author}{Lisa \surnamestart Torrey\surnameend} \&
	\bibinfo{author}{Matthew \surnamestart Taylor\surnameend}
	(\bibinfo{year}{2013}): \emph{\bibinfo{title}{Teaching on a budget: Agents
			advising agents in reinforcement learning}}.
	\newblock In: {\sl \bibinfo{booktitle}{International Conference on Autonomous
			Agents and Multi-Agent Systems (AAMAS)}}, pp. \bibinfo{pages}{1053--1060}.
	
	\bibitemdeclare{inproceedings}{trevizan2007planning}
	\bibitem{trevizan2007planning}
	\bibinfo{author}{Felipe~W. \surnamestart Trevizan\surnameend},
	\bibinfo{author}{Fabio~Gagliardi \surnamestart Cozman\surnameend} \&
	\bibinfo{author}{Leliane~Nunes \surnamestart de~Barros\surnameend}
	(\bibinfo{year}{2007}): \emph{\bibinfo{title}{Planning under risk and
			knightian uncertainty}}.
	\newblock In: {\sl \bibinfo{booktitle}{International Joint Conference on
			Artificial Intelligence (IJCAI)}}, \bibinfo{volume}{2007}, pp.
	\bibinfo{pages}{2023--2028}.
	
	\bibitemdeclare{inproceedings}{walsh2010integrating}
	\bibitem{walsh2010integrating}
	\bibinfo{author}{Thomas~J. \surnamestart Walsh\surnameend},
	\bibinfo{author}{Sergiu \surnamestart Goschin\surnameend} \&
	\bibinfo{author}{Michael~L. \surnamestart Littman\surnameend}
	(\bibinfo{year}{2010}): \emph{\bibinfo{title}{Integrating sample-based
			planning and model-based reinforcement learning}}.
	\newblock In: {\sl \bibinfo{booktitle}{AAAI Conference on Artificial
			Intelligence}}, pp. \bibinfo{pages}{612--617}.
	
	\bibitemdeclare{article}{white1986parameter}
	\bibitem{white1986parameter}
	\bibinfo{author}{Chelsea~C. \surnamestart White~III\surnameend} \&
	\bibinfo{author}{Hany~K. \surnamestart El-Deib\surnameend}
	(\bibinfo{year}{1986}): \emph{\bibinfo{title}{Parameter imprecision in finite
			state, finite action dynamic programs}}.
	\newblock {\sl \bibinfo{journal}{Operations Research}}
	\bibinfo{volume}{34}(\bibinfo{number}{1}), pp. \bibinfo{pages}{120--129},
	\doi{10.1287/opre.34.1.120}.
	
	\bibitemdeclare{article}{white1994markov}
	\bibitem{white1994markov}
	\bibinfo{author}{Chelsea~C. \surnamestart White~III\surnameend} \&
	\bibinfo{author}{Hany~K. \surnamestart Eldeib\surnameend}
	(\bibinfo{year}{1994}): \emph{\bibinfo{title}{Markov decision processes with
			imprecise transition probabilities}}.
	\newblock {\sl \bibinfo{journal}{Operations Research}}
	\bibinfo{volume}{42}(\bibinfo{number}{4}), pp. \bibinfo{pages}{739--749},
	\doi{10.1287/opre.42.4.739}.
	
	\bibitemdeclare{inproceedings}{williams2017information}
	\bibitem{williams2017information}
	\bibinfo{author}{Grady \surnamestart Williams\surnameend},
	\bibinfo{author}{Nolan \surnamestart Wagener\surnameend},
	\bibinfo{author}{Brian \surnamestart Goldfain\surnameend},
	\bibinfo{author}{Paul \surnamestart Drews\surnameend},
	\bibinfo{author}{James~M. \surnamestart Rehg\surnameend},
	\bibinfo{author}{Byron \surnamestart Boots\surnameend} \&
	\bibinfo{author}{Evangelos~A. \surnamestart Theodorou\surnameend}
	(\bibinfo{year}{2017}): \emph{\bibinfo{title}{Information theoretic MPC for
			model-based reinforcement learning}}.
	\newblock In: {\sl \bibinfo{booktitle}{IEEE International Conference on
			Robotics and Automation (ICRA)}}, \bibinfo{organization}{IEEE}, pp.
	\bibinfo{pages}{1714--1721}, \doi{10.1109/ICRA.2017.7989202}.
	
	\bibitemdeclare{inproceedings}{wolff2012robust}
	\bibitem{wolff2012robust}
	\bibinfo{author}{Eric~M. \surnamestart Wolff\surnameend}, \bibinfo{author}{Ufuk
		\surnamestart Topcu\surnameend} \& \bibinfo{author}{Richard~M. \surnamestart
		Murray\surnameend} (\bibinfo{year}{2012}): \emph{\bibinfo{title}{Robust
			control of uncertain Markov decision processes with temporal logic
			specifications}}.
	\newblock In: {\sl \bibinfo{booktitle}{Conference on Decision and Control
			(CDC)}}, \bibinfo{organization}{IEEE}, pp. \bibinfo{pages}{3372--3379},
	\doi{10.1109/CDC.2012.6426174}.
	
	\bibitemdeclare{conference}{wray2016hierarchical}
	\bibitem{wray2016hierarchical}
	\bibinfo{author}{Kyle~Hollins \surnamestart Wray\surnameend},
	\bibinfo{author}{Luis~Enrique \surnamestart Pineda\surnameend} \&
	\bibinfo{author}{Shlomo \surnamestart Zilberstein\surnameend}
	(\bibinfo{year}{2016}): \emph{\bibinfo{title}{Hierarchical Approach to
			Transfer of Control in Semi-Autonomous Systems}}.
	\newblock In: {\sl \bibinfo{booktitle}{International Joint Conference on
			Artificial Intelligence (IJCAI)}}, pp. \bibinfo{pages}{517--523}.
	
	\bibitemdeclare{inproceedings}{xu2009parametric}
	\bibitem{xu2009parametric}
	\bibinfo{author}{Huan \surnamestart Xu\surnameend} \& \bibinfo{author}{Shie
		\surnamestart Mannor\surnameend} (\bibinfo{year}{2009}):
	\emph{\bibinfo{title}{Parametric regret in uncertain Markov decision
			processes}}.
	\newblock In: {\sl \bibinfo{booktitle}{Conference on Decision and Control
			(CDC)}}, \bibinfo{organization}{IEEE}, pp. \bibinfo{pages}{3606--3613},
	\doi{10.1109/CDC.2009.5400796}.
	
\end{thebibliography}

\end{document}